# Artificial Intelligence (AI) in Action: Addressing the COVID-19 Pandemic with Natural Language Processing (NLP)

Qingyu Chen[a,*], Robert Leaman[b,*], Alexis Allot[c], Ling Luo[d], Chih-Hsuan Wei[e], Shankai Yan[f], Zhiyong Lu[g,†]

National Center for Biotechnology Information, National Library of Medicine,

National Institutes of Health 8600 Rockville Pike, Bethesda, MD, USA

*Equal contributions. †Corresponding author. Email: zhiyong.lu@nih.gov; Tel: 301-594-7089

a. 0000-0002-6036-1516, qingyu.chen@nih.gov

b. 0000-0003-3296-5766, robert.leaman@nih.gov

c. 0000-0002-2706-9054, alexis.allot@nih.gov

d. 0000-0002-5141-0259, ling.luo@nih.gov

e. 0000-0001-5094-7321, chih-hsuan.wei@nih.gov

f. 0000-0003-0369-4979, shankai.yan@nih.gov

g. 0000-0001-9998-916X, zhiyong.lu@nih.gov

Running title: Natural Language Processing for COVID-19

## Abstract

The COVID-19 pandemic has had a significant impact on society, both because of the serious health effects of COVID-19 and because of public health measures implemented to slow its spread. Many of these difficulties are fundamentally information needs; attempts to address these needs have caused an information overload for both researchers and the public. Natural language processing (NLP) – the branch of artificial intelligence that interprets human language – can be applied to address many of the information needs made urgent by the COVID-19 pandemic. This review surveys approximately 150 NLP studies and more than 50 systems and datasets addressing the COVID-19 pandemic. We detail work on four core NLP tasks: information retrieval, named entity recognition, literature-based discovery, and question answering. We also describe work that directly addresses aspects of the pandemic through four additional tasks: topic modeling, sentiment and emotion analysis, caseload forecasting, and misinformation detection. We conclude by discussing observable trends and remaining challenges.

## Introduction

### Background

Since the initial reports of an outbreak of "pneumonia of unknown cause" in Wuhan, Hubei province, China (1), the acute lack of knowledge surrounding COVID-19 (2) has driven intense investigations into its potential impact on society and interventions likely to reduce it. Once the cause was identified as a novel coronavirus, provisionally named 2019-nCoV (2), the focus of these investigations shifted to the essential questions surrounding its transmissibility and the prognosis of those infected. The virus was officially renamed SARS-CoV-2 (3) and spread worldwide in the following weeks as public health officials sought to identify measures to slow its advance and front-line healthcare workers urgently called for the available treatments to be tested. As disruptive public health measures were enacted and the death toll rose, researchers responded with a flood of articles addressing many aspects of COVID-19 and SARS-CoV-2 (4, 5). Yet the difficulty of identifying reliable and actionable knowledge specific to a particular context caused a sort of second epidemic: information overload (6, 7), exacerbated by the evolving understanding of the disease and a wave of article retractions from even well-respected journals (8, 9). Meanwhile, members of the public faced the severe psychological stress of changing public health measures, heavy economic impacts, and health uncertainties (10) while experiencing their own information overload through news and social media, aggravated by inconsistent messaging and deliberate misinformation campaigns (11, 12). At every step, success in the fight against COVID-19 has been driven by access to the right amount of reliable information and the willingness to act upon it.

Figure 1 summarizes the statistics of paper growth between January and November in LitCovid, a literature database keeping track of COVID-19 related papers in PubMed (13, 14). Since May, the number of papers in LitCovid has been growing at about 10,000 articles per month, accounting for over 7% of articles in PubMed. Each day, there are about 320 new articles that are related to COVID-19. Such dramatic increases significantly increase the burden of manual curation, analysis and, interpretation. It is pressing for natural language processing techniques to address the burden.

## Natural language processing: introduction, text genres, and pipelines

Natural language processing (NLP), also known as text mining, allows automated processing and analysis of unstructured texts, such as extracting information of interest and representing it in a structured format appropriate for computational analysis or applying transformations like summarization or translation that make the text more digestible for human readers (15). NLP thus enables analyses to be faster and use larger scales than manual analysis allows and has long been recognized as a way to alleviate information overload in biomedical research (15-17). Tools for identifying a relevant subset of documents in the scientific literature (information retrieval) have been particularly successful (18), though the trend has been towards more specific information within individual articles and more comprehensive results across the literature (19). NLP methods have matured significantly in recent years (20), with successful applications in a variety of applications, including literature-based knowledge discovery (21), facilitating analysis of high-throughput (gene expression/genome-wide association) data (22), and pharmacovigilance (23), and many others.

NLP methods have been applied to a wide variety of textual sources, which we categorize broadly into four domains. First, biomedical literature summarizes the findings of studies in publications, such as articles in PubMed and PMC. Second, clinical notes, found in electronic health records (EHR), describe a variety of clinically important aspects of an encounter with a patient at a particular visit, including their history, symptoms, diseases, test results and interventions. Third, social media reflects personal opinions, status, or emotions, such as posts on Twitter. Fourth, news articles contain descriptions of current events intended for the general population. While the results of research studies are typically published as a scientific article, applying NLP to text from other domains can provide answers to questions the literature does not yet address, for example, using social media and news to address public health issues (24). Due to the contrasting purposes of these source types, however, the text found in these sources exhibit dramatically different characteristics. Both clinical notes and social media, for example, exhibit much higher rates of spelling and grammar errors than the published literature or news articles. The content of clinical notes can vary widely between health care systems or even between individual hospitals. There are also significant differences between abstracts and full-text articles within the published literature, motivating differences in the NLP methods applied (25). NLP for clinical notes (EHRs), social media and news is primarily discussed in the Pandemic Applications section, below.

NLP tasks are typically addressed in a pipeline, with more fundamental tasks (such as information retrieval) performed before intermediate tasks (like named entity recognition: identifying mentions of critical concepts such as genes and diseases), which is in turn performed before high-level tasks (such as knowledge discovery) (26). Figure 2 shows an example of an NLP pipeline for question answering, illustrating the use of both information retrieval and named entity recognition as sub-components. Many of these tasks are well-studied, allowing existing NLP systems to be adapted to address their task in the context of the COVID-19 pandemic with a relatively small effort. However, the pandemic also presented several specific challenges. Many NLP methods use existing resources, such as a large dataset of related text (corpus) or a list of concepts, names, and relationships, to help provide meaning and context. Because both the virus and disease were new, there was initially very little relevant text available, and the relationships between the related concepts were largely unknown. The pandemic introduced completely new terminology (e.g., COVID-19) which updates to ontologies such as ICD-10 and MeSH provided relatively early in pandemic (27, 28) and existing NLP systems then had to be updated to include. Less conspicuously, the pandemic also caused significant changes in the frequency of existing terms (e.g. hydroxychloroquine, remdesivir), called a domain shift, which also necessitated system updates to avoid performance degradation. Even identifying text that mentions COVID-19 or SARS-CoV-2 presents a challenge: authors initially used descriptions because names were not available, the name of the virus was controversially changed from 2019-nCoV to SARS-CoV-2 (29), and a surprisingly high number of new name variations appear in the literature each week (30). At a higher level, NLP pipelines typically first address identifying the evidence and assertions from single articles, with later processes or even the human reader tasked with integrating the knowledge collected, considering its accuracy, or identifying which is most current. However, many critical aspects of the disease: its etiology, transmissibility, symptoms, mechanisms of action, prognostic indicators, disease course, fatality rate, long term effects – and especially questions about which interventions for prevention or treatment would be most effective – were addressed in the literature multiple times, with competing hypotheses, potentially contradictory evidence, and conclusions that were later revised or refuted. The useful knowledge about COVID-19 or SARS-CoV-2 is therefore often diffused across many articles. Finally, the pandemic is a worldwide concern, with different expert audiences (e.g., researchers, pharma, clinicians, public health officials) all requiring information that addresses their specific needs, but much of the same information needing to be disseminated and understood by non-experts in government, the media and the public in every language. Thus, a major need is the ability to transfer knowledge encoded using precise medical terms into information understandable by non-experts, possibly in another language.

## The organization of the review

In this review, we consider NLP work using published biomedical research as input or NLP used to gather data to inform biomedical research, searching PubMed and Google Scholar at the end of August 2020. We describe NLP tasks, use cases, and datasets for addressing information needs in the context of the COVID-19 pandemic. We first describe several traditional NLP tasks in the context of the COVID-19 pandemic, then consider several tasks specific to the pandemic. Specifically, we detail work on four core NLP tasks: information retrieval, named entity recognition, literature-based discovery, and question answering. We also describe work that directly addresses aspects of the pandemic through four additional tasks: topic modeling, sentiment and emotion analysis, caseload forecasting, and misinformation detection. For each task, we describe its scope, related datasets, methods, and discuss future work, which is also summarized in Figure 3

# Literature search and IR

## Introduction

Information retrieval is a set of techniques allowing users to easily and rapidly access relevant information contained in large text collections. With the rapid increase of unstructured textual information on the web in recent decades, this field has become increasingly significant, powering search engines such as Google and biomedical portals such as PubMed. However, recent events have made information retrieval more relevant than ever. Indeed, the outbreak of the COVID-19 pandemic resulted in an explosion of scientific literature, reaching a peak in June 2020 with more than 8% of all scientific output dedicated to COVID-19 (31). More than fifty thousand publications dedicated to the new coronavirus appeared in PubMed alone, as of October 2020. This "infodemic" (32) led to the emergence of numerous resources dedicated to helping doctors, biomedical researchers and the general public stay informed without being overwhelmed by the ever increasing number of new publications. Among the first was LitCovid (13, 14), introducing an easy to use topics-based navigation, quickly followed by many others. These resources satisfy a variety of information needs. First, they keep clinicians up to date with their peers' experiences with various treatments (33), management strategies, and diagnostic methods (34). They also allow researchers to share important information on viral mechanisms (35), leading to the development of new vaccines. Finally, they allow public officials to plan and assess the efficacity of various preventive measures (36) while offering the general public a trusted source for current information on the pandemic. Here we present a representative collection of these systems (Table 1), and describe the different tasks involved in their creation. These tasks include 1) selecting all relevant and only relevant information, 2) efficiently preprocessing and storing the information, and 3) allowing users to quickly and easily access relevant documents.

## Datasets

The first step of any literature resource is to decide on the scope of the collection of the relevant documents. This decision will impact the balance between exhaustiveness (allowing access to as many publications as possible) and focus (decrease the number of publications users need to go through daily). For example, resources such as CORD-19 (37) agglomerate publications related to several coronaviruses (SARS-CoV, MERS, SARS-CoV-2) while others such as LitCovid (13, 14) are strictly dedicated to SARS-CoV-2. In addition to the biological scope, literature portals differ in the type of publications integrated. Some, such as LitCovid (13, 14) are strictly dedicated to published literature, others such as CORD-19 (37) supplement published literature with preprints from bioRxiv and medRxiv, while others such as DOC Search (5) include news sources and clinical trials. As a way to combine both exhaustivity and quality, tools such as COVID Scholar (38) allow the user to filter results to only show published research or only publications specific to COVID-19. The creation of the COVID-19 Open Research Dataset (CORD-19) by the Allen Institute was an important milestone, as it allowed existing search engines to be applied to the COVID-19 literature, without each system needing to collect the relevant literature themselves and keep it updated. Following the release of CORD-19, several teams applied their existing search engines to CORD-19 data, including the Neural Covidex (39) and LIA (Ludwig declares war on COVID-19) (40). On the other hand, resources that do not rely on CORD-19, such as LitCovid, require a significant curation effort to identify relevant publications.

## Methods

Once all relevant publications have been collected, it is necessary to process them to optimize search and browsing. For users interested in a specific topic, such as a drug (ex: "remdesivir") or a group of patients (ex: "pediatrics"), free-text search represents a natural access point to the growing literature. Some systems such as iSearch (41), 2019nCoVR (42), or LitCovid (13, 14) use existing longstanding Lucene-based search solutions such as SOLR or ElasicSearch. Search performance is improved by simple techniques such as stemming or lemmatization, which transform each term into its root form (allowing, for example, "treatment" to match "treatments") and the use of domain-specific dictionaries of synonyms. LitCovid (13, 14), for example, uses synonyms extracted from MeSH (Medical Subject Headings), allowing a query term such as "cancer" to match documents containing terms such as "tumor." Other resources such as COVID Scholar (38), the Neural Covidex (39) or LIA (40) use custom search engines. The Neural Covidex (39), for example, employs a traditional Lucene-based search engine Anserini (43) for the document retrieval step and BM25 ranking, then re-ranks candidate documents using a modified T5-base deep learning model (44). In some tools such as DOC Search (5), the formulation of the search query is facilitated by smart autocomplete recognizing biomedical concepts such as drugs, species, anatomical parts, and genes.

In addition to search, facets offer an intuitive and easy way to explore available data. While most websites such as COVID-SEE (45) provide basic facets easily extractable from publications metadata such as journal, authors, and year of publication, more advanced faceting requires applying a named entity recognition (NER) system during the preprocessing step. LitCovid (13, 14), for example, offers faceting on the countries and chemicals mentioned in the article abstract, while DOC Search (5) includes types of clinical studies, mentioned age groups and genders, treatments, and patient outcomes. COVID-19 Intelligent Insight (14) goes even further by adding filters for coronavirus strains, proteins, genes or species. The SciSight system (46) is designed exclusively around facet-powered exploratory search, where instead of starting with a free-text query, users immediately select and manipulate facets. Publications can also be assigned high-level categories for better organization, such as the "Treatment," "Diagnosis," and

"Prevention" categories used in LitCovid or the "Prevention and mitigation measures" category in the COVID-19 World Information Aggregation system (47). The Global literature on coronavirus disease system (48), provided by the World Health Organization, also allows the user to filter results by the main subject of the publication and the type of study. The annotations necessary to provide these features can be achieved using automated systems, manual curation, or a combination of both.

The lack of test collections specific to COVID-19 in the early months made evaluation problematic, leading some researchers to express frustration with their inability to quantify the quality of the results provided by their system (39). Shared tasks such as the TREC-COVID (49) addressed this by allowing participants to apply their tools to the same set of documents and queries and then have their results evaluated manually. However, while results ranking performance is important, it is not an absolute criterion for literature portals comparison since the presence or absence of specific features, coverage of the literature, and the ease of use of the user interface might have more impact on the user experience than variations in document ranking (39).

## Applications

In addition to search and browsing capabilities, many portals offer distinct visualizations to facilitate the exploration of the COVID-19 literature. LitCovid (13, 14) displays a world map of countries mentioned in abstracts, DOC Search (5) shows a matrix of co-occurrence between various drugs and patient symptoms and outcomes, helping clinicians to quickly identify the best treatment for their patients. iSearch (41) clusters publications and represents them as a

FoamTree, while Neural Covidex(39) highlights the most pertinent passages in retrieved documents with the help of a BioBERT deep learning model (50) (Table 2 and Table S1). Resources also differ in the way they present search results. While most of the systems return the results as a list of publications, others such as LIA (40) return a list of matched sentences. Finally, in addition to interactive website access, most resources offer batch downloads and APIs for automated data retrieval. For example, iSearch (41) allows users to export search results into CSV or Excel format with custom columns, and LitCovid (13, 14) allows users to subscribe to personalized RSS feeds based on a search query.

## Summary

The dramatic increase in COVID-19 related literature led to the creation of numerous literature portals, leveraging natural language processing techniques to improve the performance of search and browsing, and allowing users fast and easy access to the vast number of COVID-19 publications at this critical time. However, many challenges remain.

First, the faster review process for coronavirus-related publications is subject to abuse, with several manuscripts quickly retracted by journals. One of the most famous examples is the controversial publication linking 5G networks to the new coronavirus (51), which was highly criticized and quickly retracted by the journal. Coronavirus-dedicated literature portals, therefore, need to not only add new publications regularly but also rapidly update or remove publications if needed. Some resources delayed marking the controversial 5G publication as "RETRACTED," resulting in public backlash online. Second, scientists are overwhelmed with the sheer quantity of new articles (52). The expedited review process led to an increasing amount of short low-value publications, mixed with high-quality reviews. Indeed, most publications have the lowest levels of evidence (53), and most publications are commentaries, with many authors simply sharing their views and opinions (53). To help users navigate the information overload, portals need to develop methods of prioritizing higher quality publications. However quality is inherently subjective, making it difficult to automate and any specific method is likely to be controversial. Finally, some journals have not made the full-text of coronavirus-related articles freely available (52), limiting the amount of literature available.

# Named entity recognition

## Introduction

Named entity recognition (NER) is a fundamental step in natural language processing to downstream text processing tasks. As one of the components in a traditional text mining pipeline, NER is usually the first step to provide semantic interpretations of the unstructured text, by locating and classifying the concept mentions. Early NER systems used rules and dictionaries (54). Later, machine learning approaches were widely applied to most bioconcepts (e.g., disease/chemical (55), gene (56)). The NER task consists of two steps. The first step is to recognize the boundaries of the concept spans in the text. The second step is to link the concept spans with the specific concept in the dictionary (e.g., NCBI Gene (57)). In general, every span would be assigned with an accession number (e.g., NCBI gene identifier) to represent the corresponding concept. The NER task incorporates multiple subtasks, including tokenization, sentence splitting, entity ambiguity, entity variation, etc. Different concept types encounter different levels of difficulty on each subtask. For example, the main challenge in recognizing chemicals is the high degree of variation in the concept names and chemical formulas, and the main challenge for gene concepts is their high degree of ambiguity due to the species variety.

## Datasets

Since May 2020, PubTator (58), a text-mining system with high quality automatic bioconcept annotations, has provided automatic name entity annotations for most relevant concepts (e.g., gene and disease) for the two major well-known COVID-19 relevant datasets (i.e., CORD-19 (37) and LitCovid (13, 14)). PubTator updates the annotations for these corpora daily for both abstracts as well as full text when available. In addition, the CORD-NER corpus (59) provides automated annotations on the CORD-19 dataset which covers 75 fine-grained entity types (includes gene, disease, and chemical). Details and links for both sets are listed in Table 2 and Table S1.

## Methods

In most cases, the methods of the two steps of the NER are highly independent, since the tasks of recognizing the mention boundary and linking the mention to the concept can be treated individually. However, though the information of the two steps can be utilized to each other. The methods for recognizing mentions in text could be summarized as below:

1. Dictionary-based: this method refers to the use of name lists or dictionary containing synonyms, with the purpose of recognizing the matches in the text.
2. Rule-based & regular expressions: this method is usually applied for name entities that follow a constant nomenclature (e.g., tmVar (60) to genomic variants), or name entities with highly frequent co-occurred strings, prefixes, or syntactic tags.
3. Machine learning (ML): this method treats the challenge as a sequence labeling problem to recognize the boundaries of the mentions. The method needs a set of documents with manual annotations to train a model. The most successful ML approach to biomedical concept recognition was the conditional random fields (CRF) (61) and had been heavily applied to many biomedical concepts, like TaggerOne (62). In the last few years, deep learning approaches have been used to optimize the NER performance. The two most common methods are biLSTM+CRF (63) and BERT (64). Unlike CRF, the biLSTM+CRF and BERT utilizes the advantage of language modeling to address the probability distribution and provides context to distinguish between words and phrases.

Unlike NER, the development of the normalization method is highly dependent on the specific characteristics of the bioconcepts, which is more difficult to address than NER. For example, species assignment is the main challenge of gene normalization, but not critical for other bioconcepts. Thus, the normalization step in COVID-19 research mostly applied existing tools (e.g., TaggerOne (62)) or dictionary lookup (e.g., COVID-19 SignSym (65)). The difficulties that cause the challenges of normalization are term variation and ambiguity. Focusing on the articles of the COVID-19 can significantly narrow down the scope of the candidate concepts to the recognized span in the text. Therefore, few studies (59, 66) normalized the spans to self-defined categories instead of specific concept identifiers. As an example in CORD-NER, cough and vomiting are both categorized as symptoms of COVID-19.

## System

To support the downstream NLP research of the COVID-19, several studies have addressed the named entity recognition specifically on the CORD-19 or LitCovid datasets. As shown in Table 2 and Table S1, Wang et. al. (59) used SciSpacy with distant supervision to extract multiple concepts, including gene, disease, and chemical concepts on the CORD-19 dataset. Colic et.al. (67) uses OGER (a dictionary-lookup approach with fuzzy matching) and BioBERT for the concept recognition in LitCovid. The COVID-19 SignSym system (65) extracts mentions of signs and symptoms from clinical text (electronic health records, EHRs), along with eight associated attributes, and normalizes them to LOINC codes using a dictionary-lookup approach. The performance of the NER is highly dependent on the evaluation method. Existing tools (e.g., TaggerOne) achieve 80-85% in F-measure, but if normalizing the spans to the self-defined categories, the performance is above 90% (65).

## Summary

Due to the labor required to create training corpora, developing a training corpus for a customized NER tagger for COVID-19 in a short period of time is challenging. Most studies preferred using existing tools (e.g., PubTator) and methods (e.g., BioBERT) or a dictionary lookup approach for named entity recognition tasks in COVID-19 relevant articles rather than developing new methods. A detailed description of these tools and methods are shown in Table 1. The performance of these tools and methods for COVID-19 still has room for improvement. A practical idea for quickly developing a customized system without manual curation is to apply distant supervision (59) to quickly expand the entity dictionary or training corpus. Another solution for creating a training corpus is to manually refine automated annotations made by existing tools. Our previous work (68) demonstrated that pre-annotated entities can significantly improve the efficiency and accuracy of manual annotation.

## Literature-Based Discovery

### Introduction

Literature-based discovery (LBD), also known as hypothesis generation, is the discovery of new knowledge based on known facts derived from the literature. Specifically, LBD is usually described as the process of connecting two pieces of already known knowledge previously regarded as unrelated (69). For example, in the Swanson ABC co-occurrence model (70), if text is found that explicitly states the knowledge that ''A is associated with B'' and ''B is associated with C,'' then the implicit knowledge of ''A may be associated with C'' is discovered. In the biomedical domain, LBD enables the discovery of implicit knowledge that may advance biomedical research, making it an important high-level task for biomedical natural language processing. The computational techniques used for LBD in the biomedical domain were recently surveyed (71). Here, we focus on LBD works addressing COVID-19, including related datasets, methods, and applications.

Datasets

The CORD-19 and LitCovid datasets, as shown in Table 2, are widely used for LBD studies on COVID-19, but these datasets do not provide annotations suitable for training LBD systems. However, CORD-ANN is a manually annotated corpus (72) of COVID-19 research created for training LBD systems. This corpus contains 500 sentences selected from CORD-19, with a total of 10,201 entity annotations and 9,444 relation annotations.

Methods

Most work on LBD adopts the following outline: concept extraction, hypothesis generation, and evaluation of results. First, biomedical entity recognition tools (e.g., PubTator (58) and CORD-NER (59)) are used to identify the entity concepts from text. Then, the discovery models are used to find new associations between the target concepts. Finally, the novel hypothesis is evaluated and analyzed. These LBD methods can be divided into two types: co-occurrence methods directly use co-occurrences in text as relationships between concepts while distributional methods first represent concepts using context vectors, and then find implicit relationships in vector space via vector operations and nearest neighbor search.

Most work on LBD for COVID-19 is based on co-occurrence. The first concept co-occurrence methods used concept co-occurrences to generate linking and target terms. For example, Pinto et al. (73) first used PubTator to annotate the disease, gene, and species concepts from 100,000 COVID-19 related papers, and then the human gene-disease co-occurrences supported by at least 4 papers were retained. They analyzed these associations and found that angiotensin-converting enzyme 2 (ACE2) was highly expressed in the lungs of patients with comorbidities associated with severe COVID-19. Tarasova et al. (74) discover 46 proteins related to both HIV-1 and COVID-19 from biomedical texts using the concept co-occurrence method. Karami (75) used frequency analysis to understand the significance of chemical and disease concepts, then co-occurrence analysis can assist in identifying relationships between entities in COVID-19 literature.

In addition to the co-occurrence methods, deep learning-based distributional embedding methods have been proposed for LBD. In these methods, the distributional embedding methods (e.g., BioWordVec (76) and SciBERT (77)) are first used to construct vector representations of terms, which can learn co-occurrence information from a large amount of text. Then the semantic similarity measures are leveraged to derive new scientific knowledge from an already existing one. For example, contextual embeddings from the SciBERT model that was pretrained on a large multi-domain corpus of scientific publications and fine-tuned on the CORD-19 dataset was leveraged to discover latent COVID-19 therapy targets in the scientific literature (78). BioWordVec was also used to mining the extensive biomedical literature for treatments to SARS that may also then be appropriate for COVID-19 (79).

## Application

To facilitate knowledge discovery from literature, some applications are developed to offer distinct visualization analyses. For example, SemViz (80) is a platform for semantic visualization of multiple types of entities and relations extracted from the CORD-19 dataset. These entities and relations can be visualized in many ways, e.g., word clouds, heat maps, graphs, etc. This system enables the discovery of novel inferences over relations in COVID-19 related data. Yeganova et al. (81) analyzed the LitCovid collection by applying state-of-the-art NER, classification, and clustering techniques. These analyses produce a comprehensive, synthesized view of COVID-19 research to facilitate knowledge discovery from literature. Further, LBD has important applications in drug repurposing, the process of finding new applications for existing drugs. The core objective is to discover drug-gene-disease interaction evidence from biomedical literature for drug repurposing. For example, Gates and Hamed (82) used NLP techniques to construct a drug-entity network from biomedical literature, then proposed a novel ranking algorithm, CovidX, to recommend existing drugs for potential repurposing. Patel et al. (83) built a disease-gene-drug tripartite network from biomedical text, and then using the network to identify potential new purposes for drugs already approved revealed six likely to treat comorbid symptoms of COVID-19 patients. A novel and comprehensive knowledge discovery framework, COVID-KG, was developed to extract fine-grained multimedia knowledge elements (entities, relations, and events) from scientific literature, and then was used to generate a comprehensive report for drug repurposing (84). Successful LBD applications can provide essential help for drug repurposing, which may enable the development of potential drugs for the cure of COVID-19.

## Summary

Most COVID-19 related work on LBD uses the co-occurrence model, but it is difficult to capture the complexity of biomedical processes with this simple model. Alternatively, deep learning-based distributional embedding methods have also been used to capture complex semantic associations, achieving better performance. However, deep learning models are limited by the high interpretability requirements, as it is essential that novel hypotheses are explainable. Enriching LBD technologies with explainable biomedical context is important and challenging. Many existing LBD methods and systems have been applied to the drug repurposing task, but obtaining reliable and convincing drug hypotheses in real application settings remains challenging.

# Question Answering

## Introduction

Question Answering (QA) is the NLP application that accepts a question as input, often in natural language, and outputs a ranked list of related answers or a summarized answer snippet (85). It is a joint task, merging techniques from IR (retrieving related documents or passages for a given question), text summarization (summarizing answers amongst related passages) and literature-based discovery (finding related entities mentioned in a question over knowledge bases).

There are two broad categories of QA systems: IR-based and knowledge-based (86). The main difference is that the former retrieves answers over free text whereas the latter is on structured databases such as ontologies and knowledge bases. To date, most QA systems on COVID-19 are IR-based. IR-based QA systems generally consist of three modules (85). The first module is question processing: the question is preprocessed or reformatted. The question type might be classified at this step as well (e.g., is the question asking about COVID-19 transmissions or treatments?). The second module is document/passage retrieval: finding related documents or passages regarding to the question. The third module is answer generation: constructing the final answers from the retrieved documents or passages.

QA is a critical tool for addressing information needs during the COVID-19 pandemic, reducing the load on health care professionals by automatically answering questions 24 hours a day and providing expert-curated FAQs to combat misinformation. QA is also a mature NLP application, allowing QA systems to be quickly applied to the pandemic. For example, the Penn Medicine School created a COVID-19 chatbot and made it publicly available within two weeks (87). In this section, we survey QA-related datasets, methods, and systems specifically addressing COVID-19.

## Datasets

Table 2 and Table S1 describe the existing COVID-19 related QA datasets. The datasets can be categorized into three groups based on the data type: (1) question-answer, where each instance is a question-answer pair, the most common QA data type, (2) question-category, where each instance is a question and its annotated question type, and (3) question-question, where each instance is a pair of questions and whether they are semantically similar. Here we describe representative datasets. For instance, CovidQA (42) consists of 124 question-article-answer triplets derived from 85 unique articles in CORD-19, covering 27 categories. Five curators provided annotations by synthesizing questions from the categories provided by the CORD-19 Kaggle task organizers, then manually identifying the relevant documents and locating answers.

In contrast, COVID-Q (88) consists of 1,690 COVID-19 related questions that are annotated into 15 broad categories and 207 detailed question classes. Multiple curators annotated the dataset in three phases. First, two curators discussed and assigned questions into categories. Second, an external curator verified and suggested category changes if required. Third, questions assigned to more than four question classes were further sampled and assigned to three Mechanical Turk workers. The majority vote was used for validation.

Another dataset of a different type is MQP (89). It consists of 3,048 question pairs collected from the medical domain (i.e., not specific to COVID-19) that are manually labeled as similar or different by medical doctors. Two doctors participated in the annotation with an agreement of over 85% on the 836 question pairs in a test set. While these question pairs are from the general medical domain. We included this dataset because the dataset creators have constructed it for the purpose of matching COVID-19 questions and already deployed a system using the trained models. Given a COVID-19 question, the system finds the most similar question in the FAQ pages from popular COVID-19 related websites and returns the answers accordingly.

## Methods and systems

We surveyed the existing COVID-19 QA systems and outlined them in Table 2 and Table S1. There are four QA search engines and two FAQ chatbots publicly available. The methods used by FAQ chatbots are relatively more straightforward. These methods use a corpus of curated question-answer pairs. Given a user question, the system finds the most similar question in its corpus (often by string matching) and returns the answer to that question if it is similar enough (90). In contrast, QA search engines work beyond question-answer pairs. As mentioned, the general pipeline to develop a QA search engine consists of question processing, document/passage retrieval, and answer generation modules. The detailed methods for each module are described below.

## Question processing

The question processing methods used in the COVID QA search engines can be categorized into three groups. First, traditional text processing methods, arguably the most common methods, are applied to user questions. Most systems use case folding (e.g., convert all the words to lower case), lemmatization (transforming a word into a valid base word, e.g., "testing" → "test"), and removing stop words (common words such as "a," "an" and "the"). The CAiRE-COVID system also applies sentence simplification, where complex sentences are transformed into several shorter and simpler sentences (91).

Second, model-driven processing methods are also used where a question needs to be processed into a compatible format for the later ranking models. This is particularly for the COVID QA search engines using BERT models. BERT models need the inputs that are processed in a specific format such that each word is mapped to an ID and special symbols are used to denote the start and end of a question.

Third, domain-specific processing methods are applied where the method is tailored to the biomedical domain. For example, COVIDASK used BEST (92), an NER tool, to annotate the biomedical entities mentioned in a user question.

### Document/passage retrieval

This module retrieves the documents/passages that are relevant to the processed questions. The methods used are essentially the same as those described in the Information Retrieval section. The existing COVID QA search engines used (1) traditional IR methods, such as BM25 in CAiRE-COVID, (2) deep learning-based methods, such as Sentence-BERT (93) in RECORD, and (3) both, such as BM25 and Sentence-BERT in CO-Search.

### Answer generation

This module produces final answers as output by identifying answer snippets – the text probably containing the answers – from the documents or passages retrieved. The existing COVID QA search engines use BERT-related models to identify the answer snippets. The shared approach is training a BERT model on a QA dataset from other domains but has many more instances than the existing COVID QA datasets. Most of them used SQuAD (94), consisting of over 100K question-answer pairs from over 500 articles in the general domain. Other datasets include HotpotQA (consisting of over 110K question-answer pairs from Wikipedia articles) and PubMedQA (95) (consisting of 1K manually annotated and over 250K automatically annotated question-answer pairs from PubMed articles). The models are trained in these datasets and then applied directly.

In addition to identify the answer snippets, CAiRE-COVID and CO-Search also automatically generate a summary from the selected answer snippets. In general, there are two types of text summarization methods: extractive summarization, where the key sentences from the original text are selected as a summary, and where the original text is significantly re-written. CAiRE-

COVID used both methods. For extractive summarization, it selects the top 3 answer snippets based on the cosine similarity between a question and the identified answer snippets using the BERT model. For abstractive summarization, similar to identifying answer snippets, it trained two deep learning models UniLM (96) and BART (97) on other larger datasets and then applied them to the COVID-19 context. In contrast, Co-Search used BERT and GPT-2 (98) for abstractive summarization.

## Summary

We have provided a detailed summary of NLP methods to tackle COVID-19 QA. It is impressive that QA systems were in production within two weeks to respond to the pandemic. The range of publicly available QA datasets and systems specific to COVID-19 represent a significant community effort.

We identify two primary challenges for further improvement. First, to date, the scale of COVID-19 QA datasets is still too limited to directly train deep learning models. The existing QA systems have directly applied the models trained from datasets of other domains. This is arguably the main bottleneck for identifying the most relevant answer snippets. A potential solution is to use a combination of manual and automatic annotations to generate weakly-supervised instances for training, similar to the PubMedQA dataset (95). Second, it lacks the evaluation of user experience on the COVID-19 QA systems. For instance, most of the COVID-19 QA systems use deep-learning-based models for searching for relevant documents/snippets and identifying answer snippets. Critically, these models take significantly more time for inference than traditional models, which is a concern for production systems. More thorough evaluations that incorporate other aspects of the user experience are needed.

## Pandemic-Oriented Applications

### Introduction

The wide variety of information needed during the COVID-19 pandemic has motivated a diverse assortment of task-specific NLP applications. The specific purpose of these applications and the methods they use vary widely, but they are broadly differentiated by whether they seek to inform biomedical or public health research and whether the information sought is specific, such as a forecast of the number of COVID-19 cases, or open-ended, such as the reasons for non-compliance with social distancing orders. These applications supply many case studies in applied NLP, ranging from crucial information provided by well-known classical approaches, through novel tasks addressed by cutting-edge NLP methods. As the COVID-19 pandemic is a worldwide concern, this section highlights research in languages besides English where possible.

### Methods

Since so many of the difficulties caused by the COVID-19 pandemic are new or poorly understood, many of the information needs are open-ended. Topic modeling is an NLP method that can provide a qualitative summary of a text dataset by identifying subjects that appear frequently; these can then be further analyzed if needed by stratifying the data, for example by time or geographic location. While topic modeling is commonly used in social media or news, one recent article applied it to the scientific literature to identify topics that have received less attention in SARS-CoV-2 research compared to the research on other coronaviruses (99). This study found that the SARS-CoV-2 literature is comparatively heavy on topics related to public health and clinical care rather than basic microbiology such as pathogenesis and transmission, suggesting that these may represent research opportunities. Topic modeling has also been used to identify useful unpublished clinical knowledge from social media posts by physicians (100). This study identified eight topics; the most common topic was actions and recommendations, followed by warnings about misleading information.

Applications of topic modeling for public health include a study that identified reasons for non-compliance with social distancing orders using a straightforward form of topic modeling based primarily on word counts (101), finding that the most common reasons include non-essential work and concerns over the mental and physical health implications of social distancing. A well-known but more sophisticated method, latent Dirichlet allocation, has been used to categorize public policy trends in India to understand the effectiveness of different policy changes (102). This study found that the aspects of public health messaging most strongly associated with behavior change are its consistency and breadth.

Many of the challenges caused by the COVID-19 pandemic are not caused directly by the SARS-CoV-2 virus and thus may be difficult to study. For example, social distancing mandates to prevent the spread of COVID-19 cause disruptions in day-to-day living, financial hardship, and isolation which can all have serious negative effects on mental health (103). A study using Twitter posts quantified stress levels in the United States over time using text classification but also used topic modeling to identify the primary causes (104). This study demonstrated a strong correlation in major US cities between increased stress levels and the number of COVID-19 cases, which was initially driven by fear of infection and widespread panic but later shifted primarily to financial concerns. Another study used latent Dirichlet allocation to extensively analyze topic trends in Reddit posts, a news aggregation and discussion website (105). This study found that both anxiety and suicidality increased substantially across the site during the pandemic, however the increase within mental health communities was particularly strong.

In addition to topic modeling, sentiment analysis has also been used to identify emotions and understand public opinion surveillance (106). This study identifies the emotions elicited from the COVID-19 comments on the Reddit forum. Similar studies were also performed in different social media platforms, such as Twitter and Weibo (a Chinese social media site) (107-111). Due to the lockdown in many areas, the Internet has become a primary place to express feelings and opinions, and the number of COVID-19 related posts has increased dramatically. Table 2 and Table S1 summarize related datasets and methods on COVID-19 sentiment analysis. For instance, there were over 5 million tweets in May related to workplace and school re-opening (112). Such a massive amount of data burdens manual interpretation and leads to the development of related NLP tools. Automatically tracking the emotions and sentiments over the period and across regions would facilitate how COVID-19 impacts people's well-being and lead to more effective decision-making. (111, 113)

Many important clinical questions can be addressed by analyses that build on named entity recognition, often over a relatively limited set of concepts. For example, one study attempted to identify early symptoms indicative of COVID-19 by applying the SciBERT system to identify symptoms recorded in electronic health records (EHRs) during the week prior to a COVID-19 diagnosis (114). This study noted strong associations with anosmia/dysgeusia and fever/chills. Another study determined whether symptoms could be used to decide whether an individual should be tested for COVID-19, by applying a rule-based system to EHRs (115). This study, on the other hand, concluded that despite the strong association with anosmia and dysgeusia, which are both uncommon, initial symptoms of COVID-19 are typically non-specific, diminishing the effectiveness of symptom-based screening. The COVID-19 SignSym system identifies COVID-19 signs and symptoms in hospitalized patients using an adaptation of the clinical NLP tool CLAMP (116), but also extracts the attributes body location, severity, temporal expression, subject, condition, uncertainty, negation, and course (117). Another study improved follow-up by identifying individuals whose COVID-19-positive status is recorded in the EHR free-text narrative rather than the structured attributes, by adapting the medSpaCy system (118). Fries et al. (119), on the other hand, introduce a novel weak supervision approach for named entity recognition to identify COVID-19 symptoms in EHR records. These authors also consider the novel task of identifying exposure to COVID-19, reporting a final F1 score of 80.1%. Finally, an analysis of the symptom progression of COVID-19-positive individuals over many weeks using Reddit and Facebook groups found that symptoms persist for 90 days or more in many individuals, and often include symptoms not officially associated with COVID-19 (120).

## Applications

Effectively managing clinical resources during the pandemic requires an accurate forecast of the number of cases within a geographical area. One group used a list of keywords to identify posts related to COVID-19 on Weibo, then created a classifier to differentiate between posts that report a SARS-CoV-2 infection versus posts that merely mention COVID-19 (121). These methods are well known, but accurately predicted daily case counts earlier than the official statistics by up to 14 days. Another system demonstrated that caseload forecasts can be improved by incorporating multiple data sources (122), using the RoBERTa transformer language model to extract features from news reports, then combining these in an LSTM model with other information sources to model the caseload forecast.

Combating misinformation, whether created deliberately or by mistake, has become an important task during the pandemic. Inaccurate information can spread quickly through social media networks, and most users are ill-equipped to identify science-based misinformation, especially during a rapidly evolving crisis. However automated misinformation detection is a complex task, and we identified a wide variety of approaches. One study stopped short of detecting misinformation directly and instead attempted to predict how many times a post on the social media site Weibo will be reposted, to allow these to be prioritized for review. This method employed supervised classification, using features such as words conveying emotion (123). Another approach detected misinformation by training a language model on a large amount of text about COVID-19 from a reliable source, then using the model to calculate the perplexity of the potentially misinformative text, which is a measure of surprise (124). Misinformation will typically have a high perplexity because it uses vocabulary and phrases that differ significantly from the reliable text used to train the model. Other work trains a deep learning model to directly differentiate between reliable and unreliable assertions about COVID-19 (125). Another system identifies YouTube videos containing conspiracy theories about the origin of SARS-CoV-2 (such as being caused by 5G) by analyzing the transcript of the video using a supervised machine learning approach (126). A sophisticated approach for identifying fake news converts facts extracted from a news item into logical propositions combines this with the logical form of an ontology of reliable COVID-19 information, then evaluates the joint logical system for inconsistencies (127). Finally, CoAID is a COVID-19 misinformation dataset containing annotations of accurate and inaccurate information, extracted from both news and social media posts (128).

## Summary

The far-reaching disruptions caused by the COVID-19 pandemic have created a variety of information needs that can be addressed using natural language processing (NLP). In this section, we described applications of topic modeling, sentiment and emotion analysis, and analyses created with named entity recognition over a relatively limited set of concepts. We also described NLP studies addressing caseload forecasting and misinformation detection. The methods used in these studies vary widely in sophistication, with some critical information provided by well-understood classical methods while others demonstrated the adaptability of cutting-edge methods on novel tasks.

## Conclusion

The COVID-19 pandemic has had a widespread impact on society through increased mortality and morbidity, disruptions to everyday life, and general uncertainty. Many of these difficulties are novel in type, scale or cause and one of the primary ways to address them is with access to better information, that is, the right amount of accurate information at the point where it can be put into action. In this review we have surveyed natural language processing (NLP) research using either biomedical research as input or used to inform biomedical research. This review also covers NLP on social medial and EHR data; however they are not the primary focus and more comprehensive reviews on these domains are needed.

Considering NLP work to address the COVID-19 pandemic as a whole, we see several overall themes. First, many of the existing tasks in natural language processing can directly address information needs during the COVID-19 pandemic. Information retrieval and question answering approaches directly address information needs identified by the user; information retrieval primarily by providing documents while question answering enables both queries and results in natural language. Named entity recognition provides a foundation for semantic interpretation of text. Literature-based discovery (LBD) identifies potentially new knowledge by combining information extracted from scientific articles, often augmented with additional knowledge bases. Topic modeling identifies the most common subjects in a text while sentiment analysis quantifies positive or negative affect, with both enabling further analysis through stratification and summarization. Misinformation detection is a complex high-level NLP task that has found important direct applications during the pandemic.

Second, we see a trend towards addressing aspects of the COVID-19 pandemic through rapid adaptation of existing systems. This adaptation is achieved with a variety of methods, ranging from applying classical methods – which often require no training – to complex machine learning systems retrained with a combination of manual and automatic annotations. We observe that both extremes on this scale have advantages for rapid adaptation: simple methods such as dictionaries are easily extended, while state-of-the-art machine learning methods often require fewer manually annotated examples than, either by weak supervision or transfer learning methods such as those used for deep learning transformer models that primarily train on unlabeled data and require relatively little manually annotated data for fine-tuning.

Third, we see a strong trend towards creating datasets that address aspects of the COVID-19 pandemic. The largest of these datasets consist of text that is related to the pandemic, such as collections of scientific articles or social media posts. These are typically gathered with support from automated NLP tools, and derived datasets often provide additional automated annotations. Unsurprisingly, datasets requiring significant manual effort are less common, such as those for complex tasks like question answering and literature-based discovery or large datasets for tasks like named entity recognition or sentiment analysis.

Finally, we see many efforts to address tasks that the COVID-19 pandemic brought to the forefront or that have novel aspects. These tasks include such widely varying topics as identifying topics under-studied in the literature, analyzing long COVID cases from social media, identifying mentions of exposure to SARS-CoV-2, forecasting caseloads, providing feedback on public health initiatives, quantifying mental health effects of the pandemic, and addressing misinformation. Despite some novel aspects, these efforts build on a significant amount of existing NLP work, often by mixing classical methods and cutting-edge techniques.

However significant challenges remain. While many projects have addressed different aspects of the COVID-19 pandemic, these efforts have largely remained fragmented, though notable exceptions include high-profile resources such as CORD-19. NLP systems frequently use a pipeline architecture, so that systems that address high-level tasks typically incorporate other systems. While newer deep learning models provide end-to-end training for some tasks, these architectures remain under development for many important NLP tasks. Discovering projects that provide useful software or data that can be incorporated and reused takes time, as does characterizing and integrating the systems. While this review has provided a useful overview of much of the work to date, improved interoperability and easier methods to discover relevant software and data artifacts would be beneficial.

# Acknowledgement

This research is supported by the NIH Intramural Research Program, National Library of Medicine.

# Literature Cited

1. World Health Organization. 2020. WHO | Pneumonia of unknown cause – China. *https://www.who.int/csr/don/05-january-2020-pneumonia-of-unkown-cause-china/en/*
2. World Health Organization. 2020. Novel Coronavirus ( 2019-nCoV): situation report, 22. *https://www.who.int/docs/default-source/coronaviruse/situation-reports/20200211-sitrep-22-ncov.pdf*
3. World Health Organization. 2020. Novel Coronavirus ( 2019-nCoV): situation report, 10. *https://www.who.int/docs/default-source/coronaviruse/situation-reports/20200130-sitrep-10-ncov.pdf*
4. Stegmann J. 2020. MeSH descriptors indicate the knowledge growth in the SARS-CoV-2/COVID-19 pandemic. *arXiv:2005.06259 [cs]*
5. Nowakowska J, Sobocińska J, Lewicki M, Lemańska Ż, Rzymski P. 2020. When science goes viral: The research response during three months of the COVID-19 outbreak. *Biomedicine & Pharmacotherapy = Biomedecine & Pharmacotherapie* 129: 110451
6. Brainard J. 2020. New tools aim to tame pandemic paper tsunami. *Science* 368: 924-25
7. Valika TS, Maurrasse SE, Reichert L. 2020. A Second Pandemic? Perspective on Information Overload in the COVID-19 Era. *Otolaryngology–Head and Neck Surgery*: 0194599820935850
8. Soltani P, Patini R. 2020. Retracted COVID-19 articles: a side-effect of the hot race to publication. *Scientometrics* 125: 819-22
9. Yeo-Teh NSL, Tang BL. 2020. An alarming retraction rate for scientific publications on Coronavirus Disease 2019 (COVID-19). *Accountability in Research* 0: 1-7
10. Holmes EA, O'Connor RC, Perry VH, Tracey I, Wessely S, et al. 2020. Multidisciplinary research priorities for the COVID-19 pandemic: a call for action for mental health science. *The Lancet Psychiatry* 7: 547-60
11. Tasnim S, Hossain MM, Mazumder H. 2020. Impact of Rumors and Misinformation on COVID-19 in Social Media. *Journal of Preventive Medicine and Public Health = Yebang Uihakhoe Chi* 53: 171-74
12. Rathore FA, Farooq F. 2020. Information Overload and Infodemic in the COVID-19 Pandemic. *JPMA. The Journal of the Pakistan Medical Association* 70(Suppl 3): S162-S65
13. Chen Q, Allot A, Lu Z. 2020. Keep up with the latest coronavirus research. *Nature* 579: 193
14. Xu J, Kim S, Song M, Jeong M, Kim D, et al. 2020. Building a PubMed knowledge graph. *arXiv:2005.04308 [cs]*

15. Jensen LJ, Saric J, Bork P. 2006. Literature mining for the biologist: from information retrieval to biological discovery. *Nature Reviews Genetics* 7: 119-29
16. Rzhetsky A, Seringhaus M, Gerstein M. 2008. Seeking a New Biology through Text Mining. *Cell* 134: 9-13
17. Altman RB, Bergman CM, Blake J, Blaschke C, Cohen A, et al. 2008. Text mining for biology-the way forward: opinions from leading scientists. *Genome Biology* 9: S7
18. Fiorini N, Canese K, Starchenko G, Kireev E, Kim W, et al. 2018. Best Match: New relevance search for PubMed. *PLoS Biol* 16: e2005343
19. Markowetz F. 2017. All biology is computational biology. *PLoS Biol* 15: e2002050
20. Zhao S, Su C, Lu Z, Wang F. 2020. Recent advances in biomedical literature mining. *Briefings in Bioinformatics*: bbaa057
21. Lever J, Zhao EY, Grewal J, Jones MR, Jones SJM. 2019. CancerMine: a literature-mined resource for drivers, oncogenes and tumor suppressors in cancer. *Nat Methods* 16: 505-07
22. Huang L-C, Ross KE, Baffi TR, Drabkin H, Kochut KJ, et al. 2018. Integrative annotation and knowledge discovery of kinase post-translational modifications and cancer-associated mutations through federated protein ontologies and resources. *Scientific reports* 8: 6518
23. Sarker A, Ginn R, Nikfarjam A, O'Connor K, Smith K, et al. 2015. Utilizing social media data for pharmacovigilance: A review. *J Biomed Inform* 54: 202-12
24. Baclic O, Tunis M, Young K, Doan C, Swerdfeger H, Schonfeld J. 2020. Artificial intelligence in public health: Challenges and opportunities for public health made possible by advances in natural language processing. *Canada Communicable Disease Report* 46: 161
25. Westergaard D, Stærfeldt H-H, Tønsberg C, Jensen LJ, Brunak SJPcb. 2018. A comprehensive and quantitative comparison of text-mining in 15 million full-text articles versus their corresponding abstracts. *PLOS Computational Biology* 14: e1005962
26. Rebholz-Schuhmann D, Oellrich A, Hoehndorf R. 2012. Text-mining solutions for biomedical research: enabling integrative biology. *Nat Rev Genet* 13: 829-39
27. World Health Organization. 2020. COVID-19 coding in ICD-10. [https://www.who.int/classifications/icd/COVID-19-coding-icd10.pdf](https://www.who.int/classifications/icd/COVID-19-coding-icd10.pdf)
28. U.S. National Library of Medicine. 2020. New MeSH Supplementary Concept Record for Coronavirus Disease 2019 (COVID-19). In *NLM Technical Bulletin*
29. Jiang S, Shi Z, Shu Y, Song J, Gao GF, et al. 2020. A distinct name is needed for the new coronavirus. *Lancet* 395: 949
30. Leaman R, Lu Z. 2020. A Comprehensive Dictionary and Variability Analysis of Terms for COVID-19 and SARS-CoV-2. In *Proceedings of the 1st Workshop on NLP for COVID-19 (Part 2) at EMNLP 2020*. Online: Association for Computational Linguistics
31. Odone A, Signorelli C, Stuckler D, Galea S, University Vita-Salute San Raffaele C-lmwg. 2020. The first 10,000 COVID-19 papers in perspective: Are we publishing what we should be publishing? *Eur J Public Health*
32. Gazendam A, Ekhtiari S, Wong E, Madden K, Naji L, et al. 2020. The "Infodemic" of Journal Publication Associated with the Novel Coronavirus Disease. *J Bone Joint Surg Am* 102: e64
33. Md Insiat Islam R. 2020. Current Drugs with Potential for Treatment of COVID-19: A Literature Review. *J Pharm Pharm Sci* 23: 58-64
34. Siordia JA, Jr. 2020. Epidemiology and clinical features of COVID-19: A review of current literature. *J Clin Virol* 127: 104357
35. Srivastava S, Verma S, Kamthania M, Kaur R, Badyal RK, et al. 2020. Structural Basis for Designing Multiepitope Vaccines Against COVID-19 Infection: In Silico Vaccine Design and Validation. *JMIR Bioinform Biotech* 1: e19371

36. Keeling MJ, Hollingsworth TD, Read JM. 2020. Efficacy of contact tracing for the containment of the 2019 novel coronavirus (COVID-19). *J Epidemiol Community Health* 74: 861-66
37. Wang LL, Lo K, Chandrasekhar Y, Reas R, Yang J, et al. 2020. CORD-19: The Covid-19 Open Research Dataset. In *Proceedings of the 1st Workshop on NLP for COVID-19 at ACL 2020*. Online: Association for Computational Linguistics
38. Trewartha A, Dagdelen J, Huo H, Cruse K, Wang Z, et al. 2020. COVIDScholar: An automated COVID-19 research aggregation and analysis platform. *arXiv preprint arXiv:2012.03891*
39. Zhang E, Gupta N, Nogueira R, Cho K, Lin J. 2020. Rapidly deploying a neural search engine for the covid-19 open research dataset: Preliminary thoughts and lessons learned. *arXiv preprint arXiv:2004.05125*
40. Ludwig. 2020. LIA. *https://covid19.ludwig.guru/*
41. Analysis NOoP. 2020. NIH OPA iSearch COVID-19 Portfolio. *https://icite.od.nih.gov/covid19/search/*
42. Zhao WM, Song SH, Chen ML, Zou D, Ma LN, et al. 2020. The 2019 novel coronavirus resource. *Yi Chuan* 42: 212-21
43. Yang P, Fang H, Lin J. 2017. Anserini: Enabling the Use of Lucene for Information Retrieval Research. In *Proceedings of the 40th International ACM SIGIR Conference on Research and Development in Information Retrieval*, pp. 1253–56. Shinjuku, Tokyo, Japan: Association for Computing Machinery
44. Raffel C, Shazeer N, Roberts A, Lee K, Narang S, et al. 2019. Exploring the limits of transfer learning with a unified text-to-text transformer. *arXiv preprint arXiv:1910.10683*
45. Verspoor K, Šuster S, Otmakhova Y, Mendis S, Zhai Z, et al. 2020. COVID-SEE: Scientific Evidence Explorer for COVID-19 related research. *arXiv preprint arXiv:.07880*
46. Hope T, Portenoy J, Vasan K, Borchardt J, Horvitz E, et al. 2020. SciSight: Combining faceted navigation and research group detection for COVID-19 exploratory scientific search. *arXiv preprint arXiv:2005.12668*
47. Aizawa A, Bergeron F, Chen J, Cheng F, Hayashi K, et al. 2020. A System for Worldwide COVID-19 Information Aggregation. *arXiv preprint arXiv:.01523*
48. World Health Organization. 2020. COVID-19: global literature on coronavirus disease. *https://www.who.int/emergencies/diseases/novel-coronavirus-2019/global-research-on-novel-coronavirus-2019-ncov*
49. Roberts K, Alam T, Bedrick S, Demner-Fushman D, Lo K, et al. 2020. TREC-COVID: Rationale and Structure of an Information Retrieval Shared Task for COVID-19. *J Am Med Inform Assoc*
50. Lee J, Yoon W, Kim S, Kim D, Kim S, et al. 2020. BioBERT: a pre-trained biomedical language representation model for biomedical text mining. *Bioinformatics* 36: 1234-40
51. Fioranelli M, Sepehri A, Roccia MG, Jafferany M, Olisova OY, et al. 2020. RETRACTED: 5G Technology and induction of coronavirus in skin cells. *J Biol Regul Homeost Agents* 34
52. Brainard J. 2020. Scientists are drowning in COVID-19 papers. Can new tools keep them afloat. *Science*
53. Chan J, Oo S, Chor CYT, Yim D, Chan JSK, Harky A. 2020. COVID-19 and literature evidence: should we publish anything and everything? *Acta bio-medica: Atenei Parmensis* 91: e2020020
54. Goulart RRV, de Lima VLS, Xavier CC. 2011. A systematic review of named entity recognition in biomedical texts. *Journal of the Brazilian Computer Society* 17: 103-16
55. Li J, Sun Y, Johnson RJ, Sciaky D, Wei C-H, et al. 2016. BioCreative V CDR task corpus: a resource for chemical disease relation extraction. *Database* 2016
56. Lu Z, Kao H-Y, Wei C-H, Huang M, Liu J, et al. 2011. The gene normalization task in BioCreative III. *BMC bioinformatics* 12: S2

57. Brown GR, Hem V, Katz KS, Ovetsky M, Wallin C, et al. 2015. Gene: a gene-centered information resource at NCBI. *Nucleic acids research* 43: D36-D42
58. Wei C-H, Allot A, Leaman R, Lu Z. 2019. PubTator central: automated concept annotation for biomedical full text articles. *Nucleic acids research* 47: W587-W93
59. Wang X, Song X, Li B, Guan Y, Han J. 2020. Comprehensive Named Entity Recognition on CORD-19 with Distant or Weak Supervision. *arXiv:2003.12218 [cs]*
60. Wei C-H, Harris BR, Kao H-Y, Lu Z. 2013. tmVar: A text mining approach for extracting sequence variants in biomedical literature. *Bioinformatics* 29: 1433-39
61. Lafferty J, McCallum A, Pereira FC. 2001. Conditional random fields: Probabilistic models for segmenting and labeling sequence data. In *Proceedings of the 18th International Conference on Machine Learning 2001 (ICML 2001)*, pp. 282-89
62. Leaman R, Lu Z. 2016. TaggerOne: joint named entity recognition and normalization with semi-Markov Models. *Bioinformatics* 32: 2839-46
63. Huang Z, Xu W, Yu K. 2015. Bidirectional LSTM-CRF models for sequence tagging. *arXiv preprint arXiv:1508.01991*
64. Devlin J, Chang M-W, Lee K, Toutanova K. 2018. Bert: Pre-training of deep bidirectional transformers for language understanding. *arXiv preprint arXiv:1810.04805*
65. Wang J, Pham HA, Manion F, Rouhizadeh M, Zhang Y. 2020. COVID-19 SignSym: A fast adaptation of general clinical NLP tools to identify and normalize COVID-19 signs and symptoms to OMOP common data model. *arXiv preprint arXiv:2007.10286*
66. Dong X, Li J, Soysal E, Bian J, DuVall SL, et al. 2020. COVID-19 TestNorm-A tool to normalize COVID-19 testing names to LOINC codes. *Journal of the American Medical Informatics Association*
67. Colic N, Furrer L, Rinaldi F. 2020. Annotating the Pandemic: Named Entity Recognition and Normalisation in COVID-19 Literature. In *ACL 2020*
68. Wei C-H, Harris BR, Li D, Berardini TZ, Huala E, et al. 2012. Accelerating literature curation with text-mining tools: a case study of using PubTator to curate genes in PubMed abstracts. *Database* 2012
69. Bekhuis T. 2006. Conceptual biology, hypothesis discovery, and text mining: Swanson's legacy. *J Biomedical digital libraries* 3: 2
70. Swanson DR, Smalheiser NR. 1997. An interactive system for finding complementary literatures: a stimulus to scientific discovery. *Artificial intelligence* 91: 183-203
71. Gopalakrishnan V, Jha K, Jin W, Zhang A. 2019. A survey on literature based discovery approaches in biomedical domain. *Journal of biomedical informatics* 93: 103141
72. Piad-Morffis A, Estevez-Velarde S, Estevanell-Valladares EL, Gutiérrez Y, Montoyo A, et al. 2020. Knowledge Discovery in COVID-19 Research Literature. In *Proceedings of the 1st Workshop on NLP for COVID-19 (Part 2) at EMNLP 2020*. Online: Association for Computational Linguistics
73. Pinto BG, Oliveira AE, Singh Y, Jimenez L, Gonçalves ANA, et al. 2020. ACE2 expression is increased in the lungs of patients with comorbidities associated with severe COVID-19. *MedRxiv*
74. Tarasova O, Ivanov S, Filimonov DA, Poroikov V. 2020. Data and Text Mining Help Identify Key Proteins Involved in the Molecular Mechanisms Shared by SARS-CoV-2 and HIV-1. *Molecules* 25: 2944
75. Karami A. 2020. Investigating Diseases and Chemicals in COVID-19 Literature with Text Mining.
76. Zhang Y, Chen Q, Yang Z, Lin H, Lu Z. 2019. BioWordVec, improving biomedical word embeddings with subword information and MeSH. *Scientific data* 6: 1-9
77. Beltagy I, Cohan A, Lo K. 2019. Scibert: Pretrained contextualized embeddings for scientific text. *arXiv preprint arXiv:.10676*

78. Martinc M, Škrlj B, Pirkmajer S, Lavrač N, Cestnik B, et al. 2020. COVID-19 therapy target discovery with context-aware literature mining. *arXiv preprint arXiv:.15681*
79. Kuusisto F, Page D, Stewart R. 2020. Word embedding mining for SARS-CoV-2 and COVID-19 drug repurposing. *F1000Research* 9: 585
80. Tu J, Verhagen M, Cochran B, Pustejovsky J. 2020. Exploration and Discovery of the COVID-19 Literature through Semantic Visualization. *arXiv preprint arXiv:.01800*
81. Yeganova L, Islamaj R, Chen Q, Leaman R, Allot A, et al. 2020. Navigating the landscape of COVID-19 research through literature analysis: A bird's eye view. *arXiv preprint arXiv:.03397*
82. Gates LE, Hamed AA. 2020. The Anatomy of the SARS-CoV-2 Biomedical Literature: Introducing the CovidX Network Algorithm for Drug Repurposing Recommendation. *Journal of medical Internet research* 22: e21169
83. Patel JC, Tulswani R, Khurana P, Sharma YK, Ganju L, et al. 2020. Identification of pulmonary comorbid diseases network based repurposing effective drugs for COVID-19.
84. Wang Q, Li M, Wang X, Parulian N, Han G, et al. 2020. COVID-19 Literature Knowledge Graph Construction and Drug Repurposing Report Generation. *arXiv preprint arXiv:.00576*
85. Jurafsky D, Martin JH. 2009. *Speech and language processing*: Pearson Education. 988 pp.
86. Athenikos SJ, Han H. 2010. Biomedical question answering: A survey. *Computer methods programs in biomedicine* 99: 1-24
87. Volpp Kevin G. 2020. Asked and Answered: Building a Chatbot to Address Covid-19-Related Concerns. *NEJM Catalyst Innovations in Care Delivery*
88. Wei J, Huang C, Vosoughi S, Wei J. 2020. What Are People Asking About COVID-19? A Question Classification Dataset. *arXiv preprint arXiv:.12522*
89. McCreery CH, Katariya N, Kannan A, Chablani M, Amatriain X. 2020. Effective Transfer Learning for Identifying Similar Questions: Matching User Questions to COVID-19 FAQs. In *Proceedings of the 26th ACM SIGKDD International Conference on Knowledge Discovery & Data Mining*, pp. 3458-65
90. Li Y, Grandison T, Silveyra P, Douraghy A, Guan X, et al. 2020. Jennifer for COVID-19: An NLP-Powered Chatbot Built for the People and by the People to Combat Misinformation.
91. Narayan S, Gardent C, Cohen SB, Shimorina A. 2017. Split and rephrase. *arXiv preprint arXiv:.06971*
92. Lee S, Kim D, Lee K, Choi J, Kim S, et al. 2016. BEST: next-generation biomedical entity search tool for knowledge discovery from biomedical literature. *PloS one* 11: e0164680
93. Reimers N, Gurevych I. 2019. Sentence-bert: Sentence embeddings using siamese bert-networks. *arXiv preprint arXiv:.10084*
94. Rajpurkar P, Zhang J, Lopyrev K, Liang P. 2016. Squad: 100,000+ questions for machine comprehension of text. *arXiv preprint arXiv:.05250*
95. Jin Q, Dhingra B, Liu Z, Cohen WW, Lu X. 2019. PubMedQA: A Dataset for Biomedical Research Question Answering. *arXiv preprint arXiv:1909.06146*
96. Dong L, Yang N, Wang W, Wei F, Liu X, et al. 2019. Unified language model pre-training for natural language understanding and generation. In *Advances in Neural Information Processing Systems*, pp. 13063-75
97. Lewis M, Liu Y, Goyal N, Ghazvininejad M, Mohamed A, et al. 2019. Bart: Denoising sequence-to-sequence pre-training for natural language generation, translation, and comprehension. *arXiv preprint arXiv:.13461*
98. Esteva A, Kale A, Paulus R, Hashimoto K, Yin W, et al. 2020. CO-Search: COVID-19 Information Retrieval with Semantic Search, Question Answering, and Abstractive Summarization.
99. Doanvo A, Qian X, Ramjee D, Piontkivska H, Desai A, Majumder M. 2020. Machine Learning Maps Research Needs in COVID-19 Literature. *Patterns (N Y)*: 100123

100. Wahbeh A, Nasralah T, Al-Ramahi M, El-Gayar O. 2020. Mining Physicians' Opinions on Social Media to Obtain Insights Into COVID-19: Mixed Methods Analysis. *JMIR public health and surveillance* 6: e19276
101. Moore RC, Lee A, Hancock JT, Halley M, Linos E. 2020. Experience with Social Distancing Early in the COVID-19 Pandemic in the United States: Implications for Public Health Messaging. *medRxiv: The Preprint Server for Health Sciences*
102. Debnath R, Bardhan R. 2020. India nudges to contain COVID-19 pandemic: A reactive public policy analysis using machine-learning based topic modelling. *PLOS ONE* 15: e0238972
103. Holmes EA, O'Connor RC, Perry VH, Tracey I, Wessely S, et al. 2020. Multidisciplinary research priorities for the COVID-19 pandemic: a call for action for mental health science. *Lancet Psychiatry* 7: 547-60
104. Li D, Chaudhary H, Zhang Z. 2020. Modeling Spatiotemporal Pattern of Depressive Symptoms Caused by COVID-19 Using Social Media Data Mining. *International Journal of Environmental Research and Public Health* 17
105. Low DM, Rumker L, Talker T, Torous J, Cecchi G, Ghosh SS. 2020. Natural Language Processing Reveals Vulnerable Mental Health Support Groups and Heightened Health Anxiety on Reddit during COVID-19: An Observational Study. *J Med Internet Res*
106. Jelodar H, Wang Y, Orji R, Huang H. 2020. Deep sentiment classification and topic discovery on novel coronavirus or covid-19 online discussions: Nlp using lstm recurrent neural network approach. *arXiv preprint arXiv:2004.11695*
107. Aslam F, Awan TM, Syed JH, Kashif A, Parveen M. 2020. Sentiments and emotions evoked by news headlines of coronavirus disease (COVID-19) outbreak. *Humanities and Social Sciences Communications* 7: 23-23
108. de Las Heras-Pedrosa C, Sánchez-Núñez P, Peláez JI. 2020. Sentiment Analysis and Emotion Understanding during the COVID-19 Pandemic in Spain and Its Impact on Digital Ecosystems. *International Journal of Environmental Research and Public Health* 17: 5542-42
109. Drias HH, Drias Y. 2020. Mining Twitter Data on COVID-19 for Sentiment analysis and frequent patterns Discovery. *medRxiv*: 2020.05.08.20090464-2020.05.08.64
110. Samuel J, Ali GG, Rahman M, Esawi E, Samuel Y. 2020. Covid-19 public sentiment insights and machine learning for tweets classification. *Information* 11: 314-14
111. Zhou J, Yang S, Xiao C, Chen F. 2020. Examination of community sentiment dynamics due to covid-19 pandemic: a case study from Australia. *arXiv preprint arXiv:2006.12185*
112. Ahmed ME, Rabin MRI, Chowdhury FN. 2020. COVID-19: Social Media Sentiment Analysis on Reopening. *arXiv preprint arXiv:.00804*
113. Han X, Wang J, Zhang M, Wang X. 2020. Using Social Media to Mine and Analyze Public Opinion Related to COVID-19 in China. *International Journal of Environmental Research and Public Health* 17: 2788
114. Wagner T, Shweta F, Murugadoss K, Awasthi S, Venkatakrishnan AJ, et al. 2020. Augmented curation of clinical notes from a massive EHR system reveals symptoms of impending COVID-19 diagnosis. *Elife* 9
115. Callahan A, Steinberg E, Fries JA, Gombar S, Patel B, et al. 2020. Estimating the efficacy of symptom-based screening for COVID-19. *NPJ Digit Med* 3: 95
116. Soysal E, Wang J, Jiang M, Wu Y, Pakhomov S, et al. 2018. CLAMP - a toolkit for efficiently building customized clinical natural language processing pipelines. *J Am Med Inform Assoc* 25: 331-36
117. Wang J, Pham HA, Manion F, Rouhizadeh M, Zhang Y. 2020. COVID-19 SignSym: A fast adaptation of general clinical NLP tools to identify and normalize COVID-19 signs and symptoms to OMOP common data model. *ArXiv*

118. Chapman AB, Peterson KS, Turano A, Box TL, Wallace KS, Jones M. 2020. A Natural Language Processing System for National COVID-19 Surveillance in the US Department of Veterans Affairs. In *Proceedings of the 1st Workshop on NLP for COVID-19 at ACL 2020*. Online: Association for Computational Linguistics
119. Fries JA, Steinberg E, Khattar S, Fleming SL, Posada J, et al. 2020. Trove: Ontology-driven weak supervision for medical entity classification. *arXiv preprint arXiv:2008.01972*
120. Picone M, Inoue S, DeFelice C, Naujokas MF, Sinrod J, et al. 2020. Social Listening as a Rapid Approach to Collecting and Analyzing COVID-19 Symptoms and Disease Natural Histories Reported by Large Numbers of Individuals. *Popul Health Manag*
121. Shen C, Chen A, Luo C, Zhang J, Feng B, Liao W. 2020. Using Reports of Symptoms and Diagnoses on Social Media to Predict COVID-19 Case Counts in Mainland China: Observational Infoveillance Study. *J Med Internet Res* 22: e19421
122. Zheng N, Du S, Wang J, Zhang H, Cui W, et al. 2020. Predicting COVID-19 in China Using Hybrid AI Model. *IEEE Trans Cybern* 50: 2891-904
123. Li L, Zhang Q, Wang X, Zhang J, Wang T, et al. 2020. Characterizing the Propagation of Situational Information in Social Media During COVID-19 Epidemic: A Case Study on Weibo. *IEEE Transactions on Computational Social Systems* 7: 556-62
124. Lee N, Bang Y, Madotto A, Fung P. 2020. Misinformation Has High Perplexity. *arXiv:2006.04666 [cs]*
125. Elhadad MK, Li KF, Gebali F. 2021. An Ensemble Deep Learning Technique to Detect COVID-19 Misleading Information. ed. L Barolli, KF Li, T Enokido, M Takizawa, pp. 163-75: Springer International Publishing
126. Serrano JCM, Papakyriakopoulos O, Hegelich S. 2020. NLP-based Feature Extraction for the Detection of COVID-19 Misinformation Videos on YouTube. In *Proceedings of the 1st Workshop on NLP for COVID-19 at ACL 2020*. Online: Association for Computational Linguistics
127. Groza A. 2020. Detecting fake news for the new coronavirus by reasoning on the Covid-19 ontology. *arXiv:2004.12330 [cs]*
128. Cui L, Lee D. 2020. CoAID: COVID-19 Healthcare Misinformation Dataset. *arXiv preprint arXiv:2006.00885*